\documentclass{article}
\usepackage{spconf,amsmath,epsfig}
\usepackage{epstopdf}
\usepackage{color}
\usepackage{multirow}
\usepackage{caption}
\usepackage{subfig}
\usepackage{setspace}
\usepackage{fancyhdr}
\begin{document}\sloppy

% Example definitions.
% --------------------
\def\x{{\mathbf x}}
\def\L{{\cal L}}

% Title.
% ------
\title{Distilling with Residual Network for Single Image Super Resolution}
%
% Single address.
% ---------------
\name{Xiaopeng~Sun, Wen~Lu, Rui~Wang, Furui~Bai\thanks{This research was supported in part by the National Natural Science Foundation of China (Grant Nos. 61432014, 61871311, 61876146), the National Key Research and Development Program of China (Grant No. 2016QY01W0200), the Key Industrial Innovation Chain Project in Industrial Domain (Grant No. 2016KTZDGY04-02), National High-Level Talents Special Support Program (Leading Talent of Technological Innovation of Ten-Thousands Talents Program)(Grant No. CS31117200001).}}
\address{ School of Electronic Engineering, Xidian University, Xi'an, China\\
xpsun@stu.xidian.edu.cn, luwen@mail.xidian.edu.cn, \{wangrui, frbai\}@stu.xidian.edu.cn}

\maketitle

\begin{abstract}
Recently, the deep convolutional neural network (CNN) has made remarkable progress in single image super resolution(SISR). However, blindly using the residual structure and dense structure to extract features from LR images, can cause the network to be bloated and difficult to train. To address these problems, we propose a simple and efficient distilling with residual network(DRN) for SISR. In detail, we propose residual distilling block(RDB) containing two branches, while one branch performs a residual operation and the other branch distills effective information. To further improve efficiency, we design residual distilling group(RDG) by stacking some RDBs and one long skip connection, which can effectively extract local features and fuse them with global features. These efficient features beneficially contribute to image reconstruction. Experiments on benchmark datasets demonstrate that our DRN is superior to the state-of-the-art methods, specifically has a better trade-off between performance and model size.
\end{abstract}
\section{Introduction}
\label{sec:intro}

The task of super-resolution(SR) is to reconstruct a high-resolution(HR) image consistent with it from a low-resolution(LR) image. The tasks of super-resolution are quite extensive, such as in the field of video surveillance, medical imaging, and target detection. However, SR is a reverse process of information loss. LR images have abundant low-frequency information but lose high-frequency information only in HR images. In order to address these problems, plenty of learning based methods have been applied to learn a mapping between HR images and LR images pairs.

Recently, Convolutional neural networks(CNN) are applied to a large number of visual tasks, including SR, which achieves better results than traditional methods. Dong et al.~\cite{SRCNN} firstly proposed SRCNN by a fully convolutional neural network, which could learn an end-to-end mapping between LR images and HR images, and made significant improvement over the conventional method (such as A+~\cite{A+}) with only three layers. Later, Kim et al.~\cite{VDSR} proposed VDSR increasing depth to 20 and made significant progress over SRCNN. Then, the DRCN Kim et al.~\cite{DRCN} proposed relieved the difficulty of deep network training by using gradient clips, skipping connections and recursive supervision. Lai et al.~\cite{LapSRN} proposed LapSRN that consisted of a deep laplacian pyramid, which reconstructed the HR image by step by step amplification. Based on ResNet~\cite{ResNet}, Lim et al.~\cite{EDSR} designed a very deep and wide network EDSR. Tai et al.~proposed MemNet~\cite{MemNet} consisted of memory blocks, but increased the computational complexity.~Hui et al. proposed IDN~\cite{IDN} reducing model computational complexity by distillation model. But IDN's distillation operation has the problem of information loss.~Based on the full integration of densenet and resnet, Yulun et al.~proposed RDN~\cite{RDN}, which achieved quite outstanding results. Later, Yulun et al.~introduced the attention mechanism to propose RCAN~\cite{RCAN}, and achieved amazing achievements. But these two models are too complicated, and the amount of parameters is huge.
\begin{figure}[t]
\captionsetup{belowskip=-8pt}
    \footnotesize
    \subfloat[Ground-Truth \protect\\ "{\em Image074}"]{
        \footnotesize
        \includegraphics[width=0.212\linewidth]{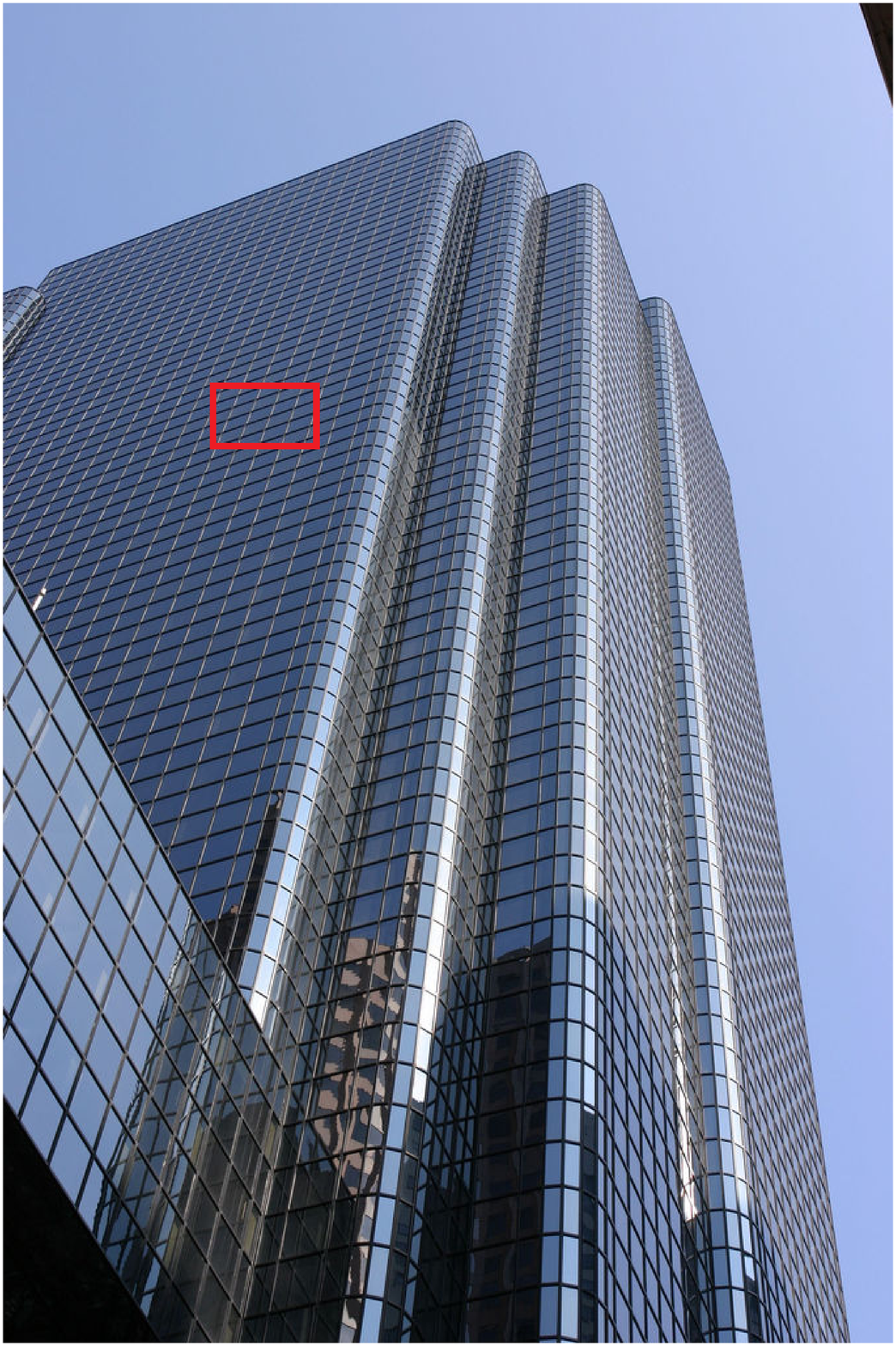}
       }
    \begin{minipage}[b]{0.85\linewidth}
        \subfloat[HR]{
            \centering
            \includegraphics[width=0.2\linewidth]{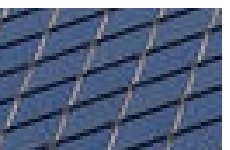}
        }
        \subfloat[Bicubic]{
            \centering
            \includegraphics[width=0.2\linewidth]{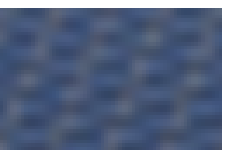}
        }
        \subfloat[LapSRN~\cite{LapSRN}]{
            \centering
            \includegraphics[width=0.2\linewidth]{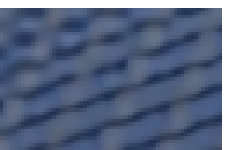}
        }
        \subfloat[MemNet~\cite{MemNet}]{
            \centering
            \includegraphics[width=0.2\linewidth]{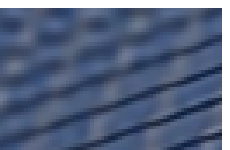}
        }\\
        \subfloat[IDN~\cite{IDN}]{
            \centering
            \includegraphics[width=0.2\linewidth]{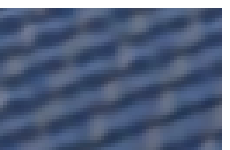}
        }
        \subfloat[SRMDNF~\cite{SRMDNF}]{
            \centering
            \includegraphics[width=0.2\linewidth]{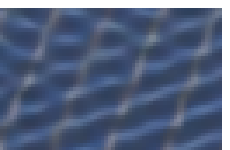}
        }
        \subfloat[MSRN~\cite{MSRN}]{
            \centering
            \includegraphics[width=0.2\linewidth]{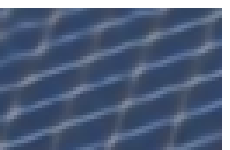}
        }
        \subfloat[DRN(ours)]{
            \centering
            \includegraphics[width=0.2\linewidth]{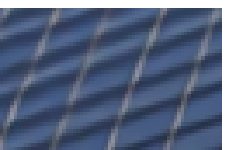}
        }
    		
        \end{minipage}

\caption{Subjective quality assessment for $4\times$ upscaling on the general image: {\em Image074} from Urban100.}
\label{Cooking}
\end{figure}
For the purpose of using different receptive fields to extract information while reducing model complexity, Li et al. proposed MSRN~\cite{MSRN} utilizing 3$\times$3 and 5$\times$5 convolutional kernel to fuse features. Although they have achieved the goal of optimizing the model, the experimental results are not outstanding enough.Their structure over-reuses features, causing the network structure to become bloated and difficult to train.

To address these problems, we propose a simple and efficient distilling with residual network for SISR. As shown in Fig\ref{fig:DPFCN}, residual distilling group(RDG) is proposed as the building module for DRN. As Fig.\ref{fig:DPFCG} shows, we stack several residual distiliing blocks(RDB) with one long skip connection(LSC) in each RDG. These long memory connections in RDBs bypass rich low-frequency information, which simplifies the flow of information. The output of one RDB can directly access each layer of the next RDB, resulting in continuous feature transfer. In addition, we introduced a convolutional layer with a 1$\times$1 kernel as feature fusion and dimensionality reduction at the last position of the RDB. The residual distilling operation is in each RDB and consists of 1$\times$1, 3$\times$3 and 1$\times$1 convolution kernels. The output of the sum of RDG and the global residual learning is sent to image reconstruction by pixelshuffle~\cite{pixelshuffle}.
\begin{figure}[t]
  \centering
  \centerline{\epsfig{figure=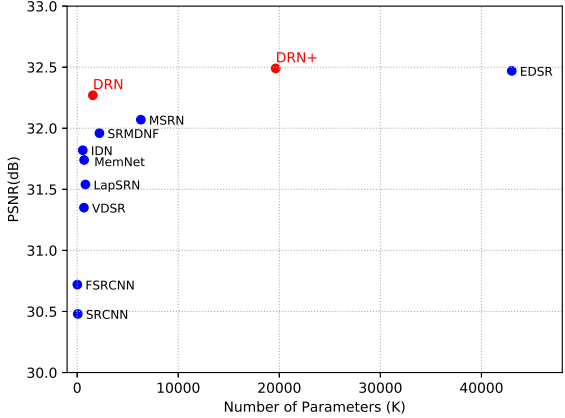,width=8.5cm}}
%  \vspace{0.5cm}
%  \centerline{(a) Result 1}\medskip
\caption{Performance and number of parameters. Results are evaluated on Set5(4$\times$). Our models have a
better trade-off between performance and model size.}
\label{fig:parameters}
\end{figure}
In summary, our main contributions are three-fold:

\begin{itemize}
	\item We propose residual distilling block(RDB), which enjoys benefits from ResNet~\cite{ResNet} and distills efficient imformation. It can fuse common feature while maintaining the ability to distill important features. Different from IDN~\cite{IDN}, our RDB using residual distilling structure, retains as much information.
	\item Our method has few network parameters and a simple network structure, which is easy to recurrent. It is a compact network with a significant trade-off between performance and model size.
	\item  We propose a simple and efficient distilling with residual network(DRN) for high-quality image SR. What's more, it is easy to understand and better than most of the state-of-the-art methods.
\end{itemize}

\section{Distilling with Residual Network}

\subsection{Network Architecture}

\begin{figure}[t]
  \centering
  \centerline{\epsfig{figure=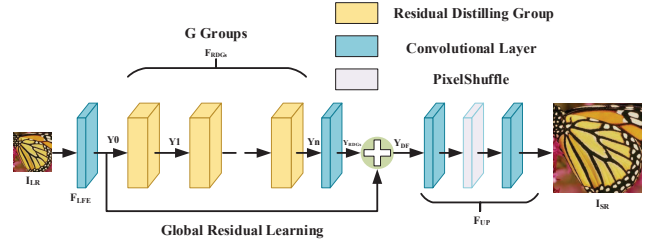,width=8.5cm}}
%  \vspace{0.5cm}
%  \centerline{(a) Result 1}\medskip
\caption{Distilling with Resiudal Network Network(DRN).}
\label{fig:DPFCN}
\end{figure}

As shown in Fig.\ref{fig:DPFCN}, the proposed DRN mainly consists three parts: low-level feature extraction(LFE), residual distilling groups(RDGs), image reconstruction(IR). Here, let's denote $I_{LR}$ and $I_{HR}$ as the input and output of DRN. As referred in ~\cite{SRGAN,EDSR,RCAN}, one convolutional layer is suitable to extract the low-level feature $Y_{0}$ from the input LR
\begin{eqnarray}
    Y_{0} = F_{LFE}(I_{LR}),
\end{eqnarray}
where $F_{LFE}$ represents convolutional function. $Y_{0}$ is then sent to the residual distilling groups and used for global residual learning. Furthermore, we can have $Y_{RDGs}$ that's the output of GDGs
\begin{eqnarray}
\begin{split}
    Y_{RDGs} = F_{RDGs}(Y_{0}),
\end{split}
\end{eqnarray}
where $F_{RDGs}$ denotes the operations of the RDGs we proposed, which contains $G$ groups.~With the deep feature information being extracted by a set of RDGs, we can further fuse the features, which contains global residual learning and $Y_{RDGs}$ . So, we have all the features extracted $Y_{DF}$,
\begin{eqnarray}
Y_{DF} = Y_{0}+Y_{RDGs}.
\end{eqnarray}
Then $Y_{DF}$ is upscaled through image reconstruction module. We can get upscaled feature $Y_{UP}$
\begin{eqnarray}
Y_{UP} = F_{UP}(Y_{DF}),
\end{eqnarray}
where $F_{UP}$ denotes the image reconstruction module. Then the upscaled feature is reconstructed by one convolution layer. In general, the overall process can be expressed as
\begin{eqnarray}
I_{SR} = F_{IR}(F_{UP}) = F_{IR}(F_{RDGs}(F_{LFE}(I_{LR}))),
\end{eqnarray}
where $F_{IR}$ and $F_{RDGs}$ denote the image reconstruction and the function of our RDGs repectively.
\subsection{Residual Distilling Group}
\begin{figure}[t]
  \centering
  \centerline{\epsfig{figure=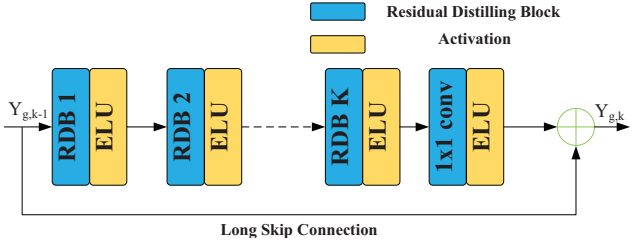,width=8.5cm}}
%  \vspace{0.5cm}
%  \centerline{(a) Result 1}\medskip
\caption{Residual Distilling Group(RDG).}
\label{fig:DPFCG}
\end{figure}

We now give more details about RDG. Through Fig.\ref{fig:DPFCG}, we can see each group contains $K$ residual distilling blocks(RDBs) and one long skip connection(LSC). Such our structure can achieve  high performance in image super resolution with a general number of convolution layers.

With all of the above, a RDG in $g$-th group is represented as
\begin{eqnarray}
Y_{g} &=& F_{g}(Y_{g-1})+Y_{g-1}\\
&=& F_{g}(F_{g-1}(\cdot \cdot \cdot F_{1}(Y_{0})\cdot \cdot \cdot ))+Y_{g-1},
\end{eqnarray}
where $F_{g}$ denotes the function of $g$-th RDG. $F_{g-1}$ and $F_{g}$ are the input and output of $g$-th RDG. Unlimited use of the residual distilling will increase the number of channels by a very large amount.~Therefore, we set a 1$\times$1 convolution with ELU~\cite{ELU} to reduce the number of channels, but it also can combine the fused distillation features together. Finally, when $g$ is $G$, we have the output of RDGs
\begin{eqnarray}
Y_{RDGs}=Y_{G},
\end{eqnarray}
where $Y_{G}$ denotes the output of $G$-th RDG.
\subsection{Resiudal Distilling Block}

The LR images have abundant low frequency information, except high frequency information the HR images only have. Therefore, we need to extract LR information and generate high frequency information. From the perspective of feature sharing of learning, ~\cite{DPN} found that the connection in residual learning is an effective way to eliminate the phenomenon of disappearing gradients in deep networks. Inspired by the recent success of by~\cite{DPN} on ImageNet, we design resiudal distilling as basic convolution in each RDB, that is, when the channel performs the residual operation, it simultaneously distills out the new channel. The channel operated by the residual operation, retains the input information as much as possible and the new distilled channel contains the useful features which is conducive to generate high frequency information. The resiudal distilling inherits the advantages of ResNet~\cite{ResNet} and distilled efficient information, to achieve an effective reuse and re-exploitation.

\begin{figure}[t]
  \centering
  \centerline{\epsfig{figure=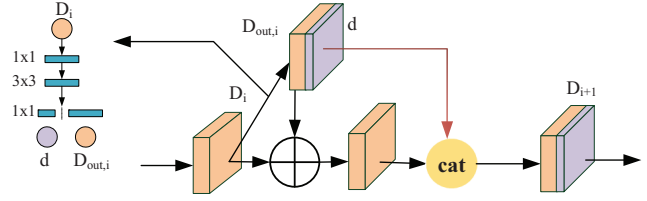,width=8.5cm}}
%  \vspace{0.5cm}
%  \centerline{(a) Result 1}\medskip
\caption{Resiudal Distilling Block(RDB).}
\label{fig:DPB}
\end{figure}

For intuitive understanding, as shown in Fig.\ref{fig:DPB}, let’s $D_{i}$ denotes the feature map dimensions of the $i$-th layers. In this way, the relationship of the convolution layers can be expressed as:
\begin{eqnarray}
D_{i+1} - D_{i} = d,
\end{eqnarray}
where $d$ denotes the channel that is distilled out between ($i+1$)-th layer and $i$-th layer. The number of $D_{i}$ dimensions perform residual operation, and $d$ dimensions perform cat operation. The whole process can be expressed as:
\begin{eqnarray}
[D_{out,i}, d] = F_{rd}(D_{i}),
\end{eqnarray}
\begin{eqnarray}
D_{i+1} = F_{concat}(D_{out,i}+D_{i}, d),
\end{eqnarray}
where [$D_{out,i},d$] are output of $D_{i}$ by convolutional function and $F_{rd}$ denotes resiudal distilling function in Fig.\ref{fig:DPB} left, and $F_{concat}$ represents concatenation operation . The dimensions of $D_{out,i}$ is same as $D_{i}$. Through this process, local residual information have been extracted by residual operation, and the net still remains a distilled path to learn new features flexibility. As we all know, high resolution to low resolution is a process of information degradation. Therefore, resiudal distilling helps the neural network ectract the useful features through potential information.

As shown in Fig.\ref{fig:DPFCG}, we stack RDBs in one RDG with one long connection. Too deep a network can cause the learned features to disappear. We design a long memory connection to allow the network reserve information about the previous block. We steak $K$ resiudal distilling blocks(RDB) in each RDG. So $Y_{g,k}$, the $k$-th resiudal distilling block in $g$-th RDG, can be expressed as
\begin{eqnarray}
Y_{g,k} =& F_{RDB,k}(D_{k})\\
=& F_{RDB,k}(F_{RDB,k-1}(...F_{RDB,1}(D_{1})...))),
\end{eqnarray}
where  $F_{RDB,K}$ represents the $k$-th RDB function. The resiudal distilling block is simple, lightweight and accurate. Finally, we can get  $Y_{g}$, the output of $g$-th RDG
\begin{eqnarray}
F_{g}(Y_{g-1}) &=& F_{p}(Y_{g,K}),\\
Y_{g} &=& F_{g}(Y_{g-1})+Y_{g-1},
\end{eqnarray}
where $F_{p}$ denotes the compression using 1$\times$1 convolution with ELU, and $Y_{g,K}$ is the output of $g$-th RDG when $k$ is $K$.
\subsection{Image Reconstruction}

As discussed in Session 2.3, the output $ Y_{0}$ and $Y_{RDGs}$ of the previous network represent global residual information and deep information respectively. Send the result $Y_{DF}$ of the two additions to the upsampling module.

There are several methods to upscaling modules, such as deconvolution layer, nearest-neighbor upsampling convolution and pixelshuffle proposed by ESPCN~\cite{pixelshuffle}. However, with the upscaling factor increasing, the network will have some uncertain training problems. The weight of the deconvolution will change with the network training. Furthermore, these methods can’t work on odd upscaling factors(e.g. x3, x5). Based on the above situation, we choose pixelshuffle as upscaling module due to the best performance.Detailed parameters of pixelshuffle are in Table  1.

\begin{figure}[t]
\captionsetup{belowskip=-8pt}
    \footnotesize
    \subfloat[Ground-Truth \protect\\ "{\em Image027}"]{
        \footnotesize
        \includegraphics[width=0.253\linewidth]{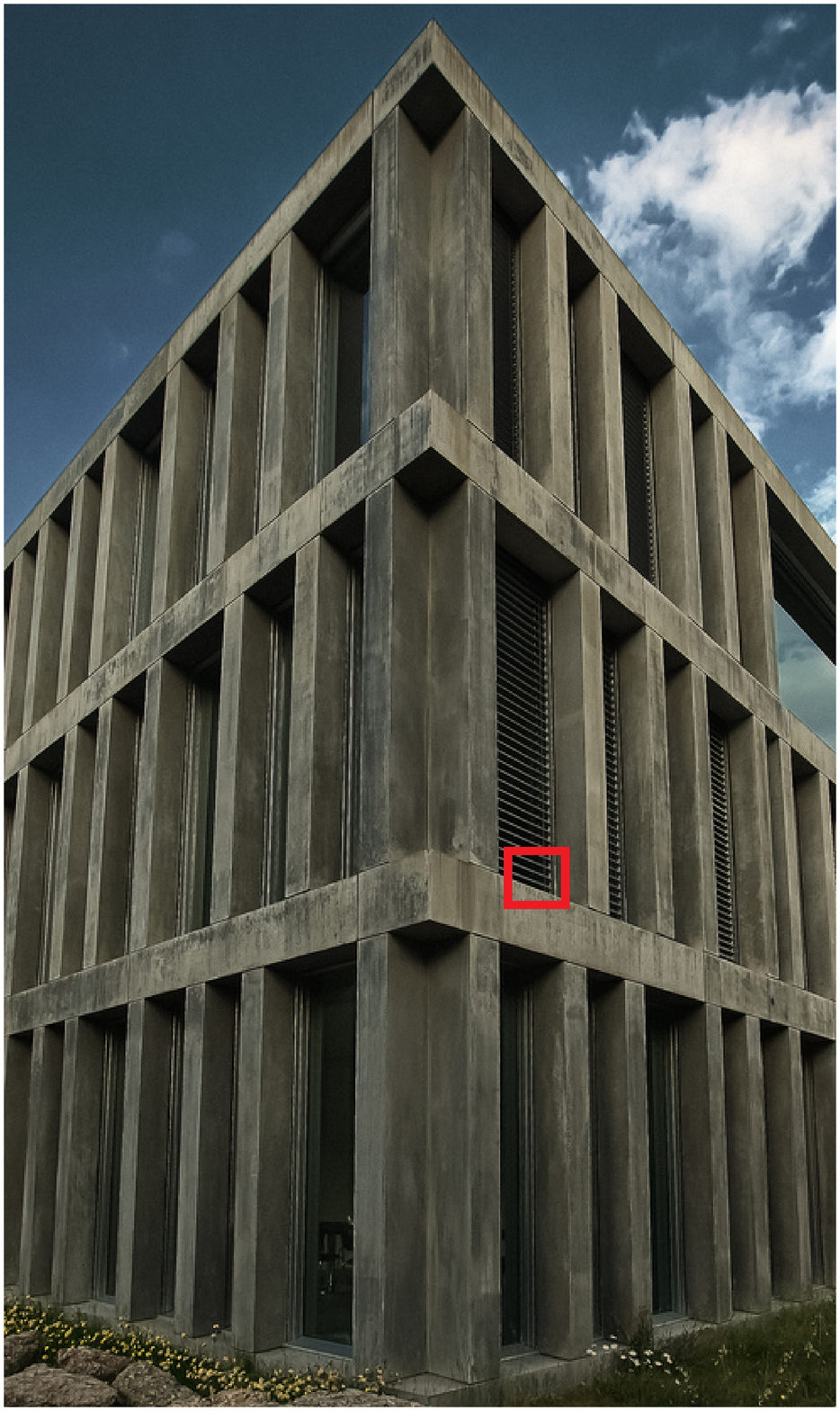}
       }
    \begin{minipage}[b]{0.85\linewidth}
        \subfloat[HR]{
            \centering
            \includegraphics[width=0.2\linewidth]{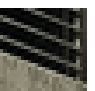}
        }
        \subfloat[Bicubic]{
            \centering
            \includegraphics[width=0.2\linewidth]{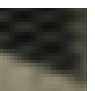}
        }
        \subfloat[LapSRN\cite{LapSRN}]{
            \centering
            \includegraphics[width=0.2\linewidth]{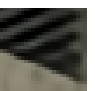}
        }
        \subfloat[MemNet\cite{MemNet}]{
            \centering
            \includegraphics[width=0.2\linewidth]{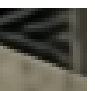}
        }\\
        \subfloat[IDN\cite{IDN}]{
            \centering
            \includegraphics[width=0.2\linewidth]{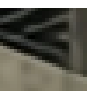}
        }
        \subfloat[SRMDNF\cite{SRMDNF}]{
            \centering
            \includegraphics[width=0.2\linewidth]{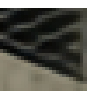}
        }
        \subfloat[MSRN\cite{MSRN}]{
            \centering
            \includegraphics[width=0.2\linewidth]{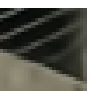}
        }
        \subfloat[DRN(ours)]{
            \centering
            \includegraphics[width=0.2\linewidth]{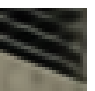}
        }
    		
        \end{minipage}

\caption{Subjective quality assessment for $3\times$ upscaling on the general image: {\em Image027} from Urban100.}
\label{fig:Image027 x3}
\end{figure}
\begin{figure}
\captionsetup{belowskip=-8pt}
    \footnotesize
    \subfloat[Ground-Truth \protect\\ "{\em ppt3}"]{
        \footnotesize
        \includegraphics[width=0.2\linewidth]{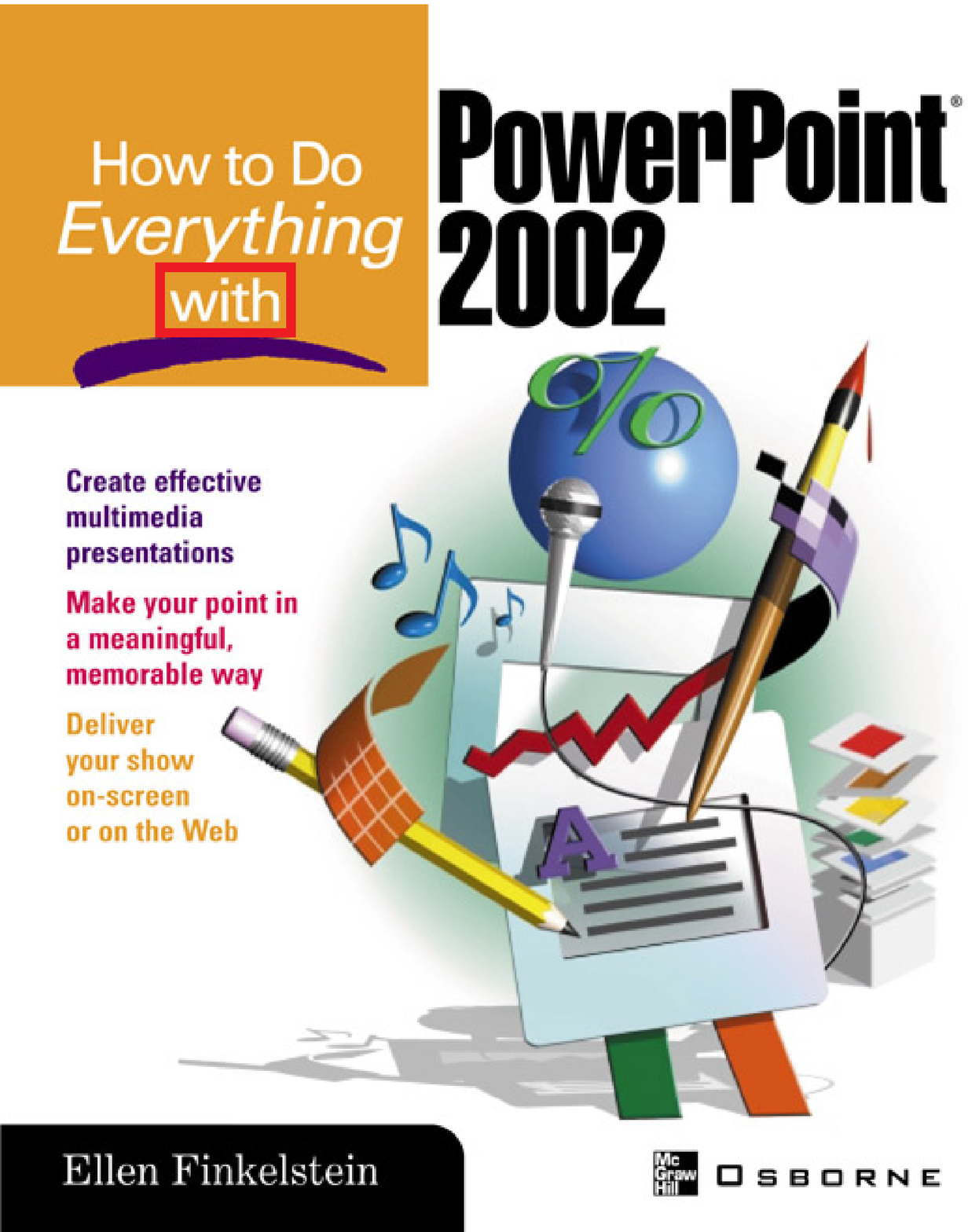}
       }
    \begin{minipage}[b]{0.85\linewidth}
        \subfloat[HR]{
            \centering
            \includegraphics[width=0.2\linewidth]{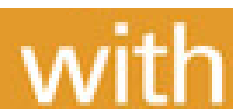}
        }
        \subfloat[Bicubic]{
            \centering
            \includegraphics[width=0.2\linewidth]{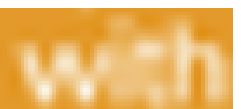}
        }
        \subfloat[LapSRN\cite{LapSRN}]{
            \centering
            \includegraphics[width=0.2\linewidth]{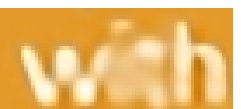}
        }
        \subfloat[MemNet\cite{MemNet}]{
            \centering
            \includegraphics[width=0.2\linewidth]{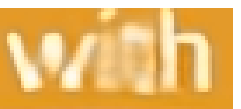}
        }\\
        \subfloat[IDN\cite{IDN}]{
            \centering
            \includegraphics[width=0.2\linewidth]{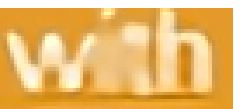}
        }
 	   \subfloat[SRMDNF\cite{SRMDNF}]{
            \centering
            \includegraphics[width=0.2\linewidth]{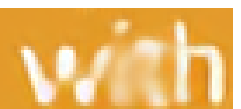}
        }
        \subfloat[MSRN\cite{MSRN}]{
            \centering
            \includegraphics[width=0.2\linewidth]{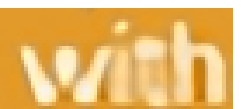}
        }
        \subfloat[DRN(ours)]{
            \centering
            \includegraphics[width=0.2\linewidth]{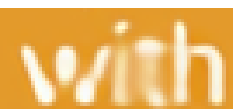}
        }   		
        \end{minipage}

\caption{Subjective quality assessment for $4\times$ upscaling on the general image: {\em ppt3} from Set14.}
\label{fig:Image044 x3}
\end{figure}
\begin{figure}[!t]
\captionsetup{belowskip=-8pt}
    \footnotesize
    \subfloat[Ground-Truth \protect\\ "{\em Image059}"]{
        \footnotesize
        \includegraphics[width=0.26\linewidth]{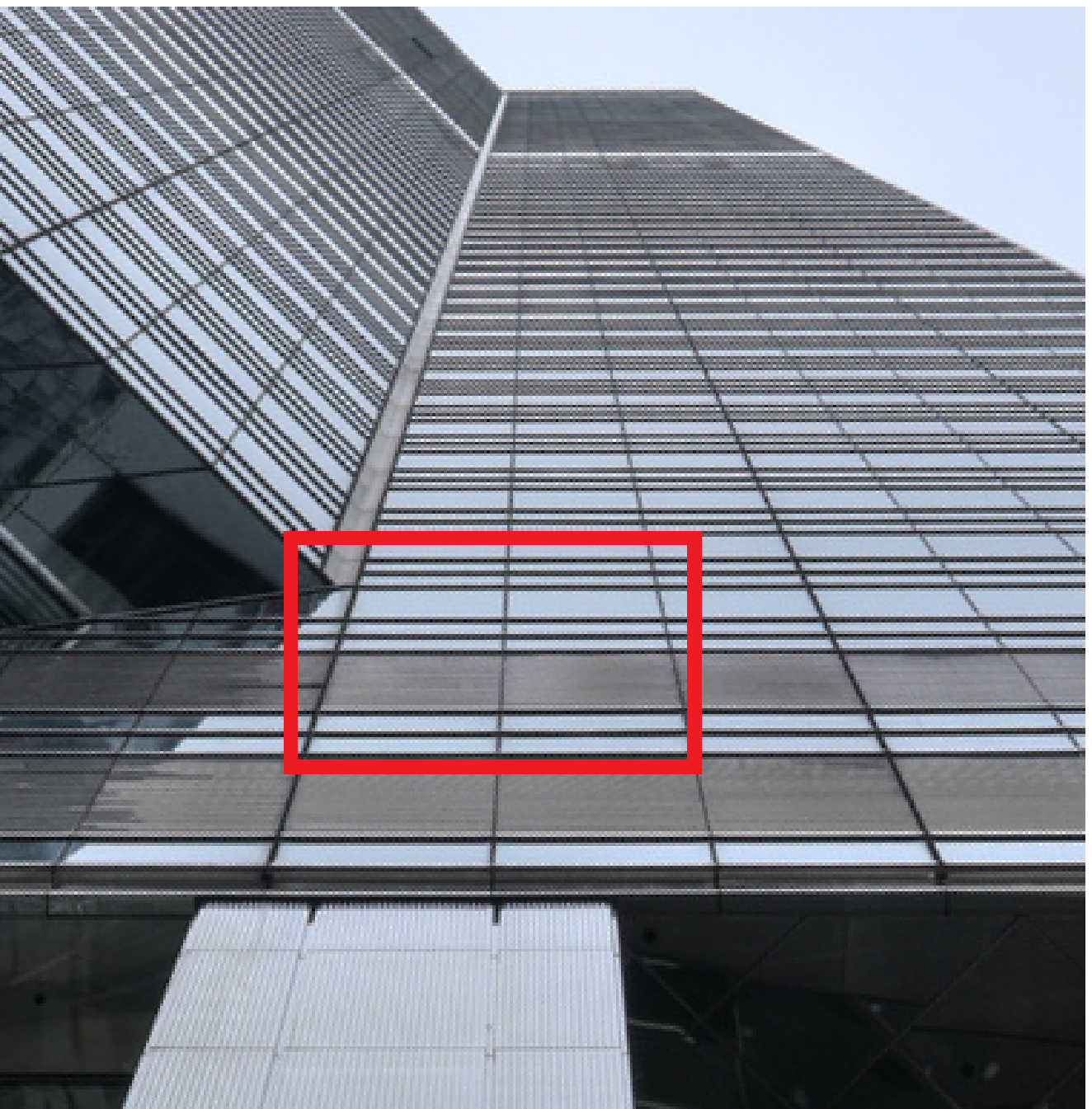}
       }
    \begin{minipage}[b]{0.85\linewidth}
        \subfloat[HR]{
            \centering
            \includegraphics[width=0.2\linewidth]{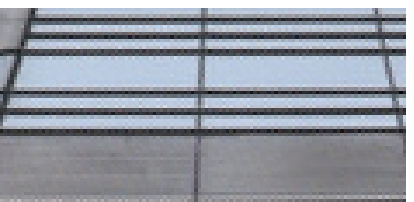}
        }
        \subfloat[Bicubic]{
            \centering
            \includegraphics[width=0.2\linewidth]{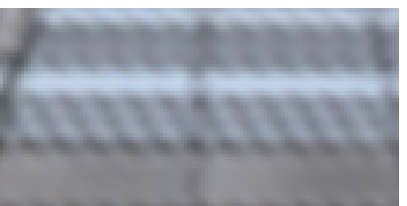}
        }
        \subfloat[LapSRN\cite{LapSRN}]{
            \centering
            \includegraphics[width=0.2\linewidth]{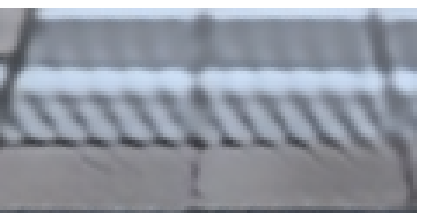}
        }
        \subfloat[MemNet\cite{MemNet}]{
            \centering
            \includegraphics[width=0.2\linewidth]{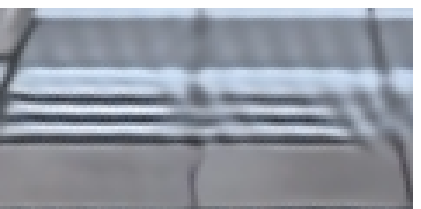}
        }\\
        \subfloat[IDN\cite{IDN}]{
            \centering
            \includegraphics[width=0.2\linewidth]{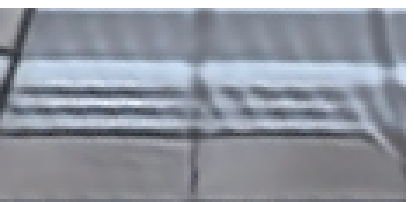}
        }
 	   \subfloat[SRMDNF\cite{SRMDNF}]{
            \centering
            \includegraphics[width=0.2\linewidth]{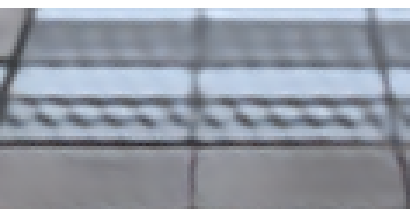}
        }
        \subfloat[MSRN\cite{MSRN}]{
            \centering
            \includegraphics[width=0.2\linewidth]{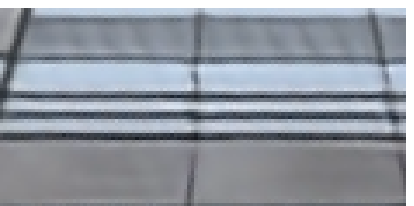}
        }
        \subfloat[DRN(ours)]{
            \centering
            \includegraphics[width=0.2\linewidth]{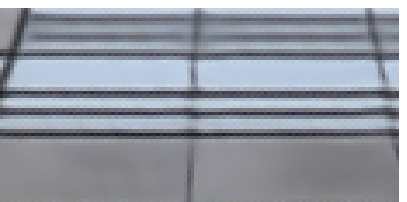}
        }   		
        \end{minipage}

\caption{Subjective quality assessment for $4\times$ upscaling on the general image: {\em Image059} from Urban100.}
\label{fig:Image059 x4}
\end{figure}

\subsection{Loss function}

There are many loss functions available in the super-resolution field, such as mean square error(MSE)~\cite{SRCNN,VDSR,FSRCNN}, mean absolute loss~\cite{LapSRN,EDSR}, perceptual and adversarial loss~\cite{SRGAN}. With the MSE loss, the neural networks generate images that are not in line with human vision~\cite{LapSRN}, so we optimize the model with MAE that is formulated as follows:
\begin{eqnarray}
l_{MAE} = \frac{1}{N} \sum_{i=1}^{N }\left \| I_{i}-\hat{I}_{i}  \right \|_{1}
\end{eqnarray}
where $N$ denotes the number of training samples in each batch. $ I_{i}$ is the reconstructed HR image. $\hat{I}_{i}$ denotes the ground truth HR image respectively. We also make a comparison of the results of using MAE and MSE respectively, as shown in Fig.\ref{fig:loss}.

\begin{table}[t]
\begin{center}
\caption{Detailed configuration information about the reconstruction structure. } \label{tab:cap}
\begin{tabular}{|c|c|c|}
  \hline
  % after \\: \hline or \cline{col1-col2} \cline{col3-col4} ...
  Laye name & Input channel & Output channel
  \\
  \hline
  conv input & C & C$\times$M$\times$M \\
  PixelShuffle($\times$M) & C$\times$M$\times$M & C \\
  conv output & C & 3\\
  \hline
\end{tabular}
\end{center}
\label{tab:pixelshuffle}
\end{table}

\begin{table}[t]
\begin{center}
\caption{Quantitative comparison of results with or without RDB on Set5(3$\times$) at 100th epoch.} \label{tab:cap}
\begin{tabular}{|c|c|c|c|}
  \hline
  % after \\: \hline or \cline{col1-col2} \cline{col3-col4} ...
  Methods & RDBs & No RDBs & DBN+
  \\
  \hline
  PSNR on Set5(3$\times$) & 33.98 & 33.83 & 34.35 \\
  \hline
\end{tabular}
\end{center}
\label{tab:compare}
\end{table}

\begin{table*}[t]
%\captionsetup{belowskip=-14pt}
\footnotesize
\centering
\caption{Benchmark results of state-of-the-art SR methods: Average PSNR/SSIM/IFC for 2$\times$, 3$\times$, and 4$\times$ upscaling. The bold figures indicate the best performance.}
\setlength{\tabcolsep}{0.35mm}{
\begin{tabular}{{|c|c|c|c|c|c|c|c|c|c|c|c|}}
\hline
 Dataset & Scale & Bicubic &A+~\cite{A+} & VDSR~\cite{VDSR} &DRCN~\cite{DRCN} &LapSRN~\cite{LapSRN} &IDN~\cite{IDN} &SRMDNF~\cite{SRMDNF} &MSRN~\cite{MSRN} &~DRN(ours) &~DRN+(ours) \\
\hline
\hline
\multirow{3}*{Set5}
	& $\times2$
    & 33.66/0.9300 & 36.60/0.9542 & 37.53/0.9583 & 37.63/0.9584 & 37.52/0.9581 & 37.83/0.9600 &37.79/0.9601 & {\color{blue}{38.08}}/0.9605 & 38.06/{\color{blue}{0.9607}} & {\color{red}{38.18}}/{\color{red}{0.9612}}\\
	& $\times3$
    & 30.39/0.8688 & 32.63/0.9085 & 33.66/0.9201 & 33.82/0.9215 & 33.82/0.9207 & 34.11/0.9253 &24.12/0.9254 & 34.38/0.9262 & {\color{blue}{34.45}}/{\color{blue}{0.9274}} & {\color{red}{34.68}}/{\color{red}{0.9293}}\\
    & $\times4$
    & 28.42/0.8104 & 30.33/0.8565 & 31.35/0.8838 & 31.53/0.8854 & 31.54/0.8852 & 31.82/0.8903 &31.96/0.8925 & 32.07/0.8903 &  {\color{blue}{32.27}}/ {\color{blue}{0.8964}} &  {\color{red}{32.49}}/ {\color{red}{0.8985}}\\
\hline
\hline
\multirow{3}*{Set14}
	& $\times2$
    & 30.24/0.8688 & 32.42/0.9059 & 33.03/0.9124 & 33.04/0.9118 & 33.08/0.9124 & 33.32/0,9159 & 33.30/0.9148 &  {\color{blue}{33.74}}/0.9170& 33.64/{\color{blue}{0.9179}} & {\color{red}{33.85}}/{\color{red}{0.9193}}\\
	& $\times3$
    & 27.55/0.7742 & 29.25/0.8194 & 29.77/0.8314 & 29.76/0.831 & 29.87/0.8325 & 29.99/0.8354 &30.04/0.8382 & {\color{blue}{30.34}}/0.8395 & 30.30/{\color{red}{0.8664}} & {\color{red}{30.57}}/{\color{blue}{0.8466}}\\
    & $\times4$
    & 26.00/0.7027 & 27.44/0.7450 & 28.01/0.7674 & 28.02/0.7670 & 28.19/0.7700 & 28.25/0.7730 & 28.35/0.7787 & 28.60/0.7751 & {\color{blue}{28.69}}/{\color{blue}{0.7839}} & {\color{red}{28.83}}/{\color{red}{0.7872}}\\
\hline
\hline
\multirow{3}*{BSDB100}
	& $\times2$
    & 29.56/0.8431 & 31.24/0.8870 & 31.90/0.8960 & 31.85/0.8942 & 31.80/0.8952 & 32.08/0.8985 & 32.05/0.8985 & {\color{blue}{32.23}}/{\color{red}{0.9013}} & {\color{blue}{32.23}}/{\color{blue}{0.9001}} & {\color{red}{32.32}}/{\color{red}{0.9013}}\\
	& $\times3$
    & 27.21/0.7385 & 26.05/0.8019 & 28.82/0.7976 & 28.80/0.7963 & 28.82/0.7980 & 28.95/0.8013 &28.97/0.8025 & 29.08/0.8554 & {\color{blue}{29.12}}/{\color{blue}{0.8055}} & {\color{red}{29.26}}/{\color{red}{0.8090}}\\
    & $\times4$
    & 25.96/0.6675 & 26.83/0.6999 & 27.29/0.7251 & 27.23/0.7232 & 27.32/0.7284 & 27.41/0.7297 &27.49/0.7337 &27.52/0.7273 & {\color{blue}{27.65}}/{\color{blue}{0.7380}} & {\color{red}{27.72}}/{\color{red}{0.7403}}\\
\hline
\hline
\multirow{3}*{Urban100}
	& $\times2$
    & 26.88/0.8403 &29.25/0.8955 & 30.76/0.9140 & 30.75/0.9133 & 30.41/0.9103 & 31.27/0.9196 & 31.33/0.9204 & {\color{blue}{32.22}}/{\color{blue}{0.9326}} & {\color{blue}{32.22}}/0.9288 & {\color{red}{32.65}}/{\color{red}{0.9329}}\\
	& $\times3$
    & 24.46/0.7349 &26.05/0.8019 & 27.14/0.8279 & 27.15/0.8276 & 27.07/0.8275 & 27.42/0.8359 & 27.57/0.8398  & 28.08/{\color{blue}{0.8554}} & {\color{blue}{28.18}}/0.8520 & {\color{red}{28.73}}/{\color{red}{0.8627}}\\
    & $\times4$
    & 23.14/0.6577 & 24.34/0.7211 & 25.18/0.7524 & 25.14/0.7510 & 25.21/0.7562 & 25.41/0.7632 & 25.68/0.7731 &26.04/0.7896 & {\color{blue}{26.26}}/{\color{blue}{0.7903}} & {\color{red}{26.54}}/{\color{red}{0.7982}}\\
\hline
\hline
\multirow{3}*{Manga109}
	& $\times2$
	& 30.82/0.9332 & 35.37/0.9663 & 37.22/0.9729 & 37.63/0.9723 & 37.27/0.9855 & 38.02/0.9749 & 38.07/0.9761 & {\color{blue}{38.82}}/{\color{red}{0.9868}} & 38.75/0.9773 & {\color{red}{38.94}}/{\color{blue}{0.9779}}\\
	& $\times3$
	& 26.96/0.8555 & 29.93/0.9089 & 32.01/0.9310 & 32.31/0.9328 & 32.21/0.9318 & 32.79/0.9391 & 33.00/0.9403 & 33.44/0.9427 & {\color{blue}{33.78}}/{\color{blue}{0.9455}} & {\color{red}{34.21}}/{\color{red}{0.9486}} \\
	& $\times4$
	& 24.91/0.7826 & 27.03/0.8439 & 28.83/0.8809 & 28.98/0.8816 & 29.09/0.8845 & 29.41/0.8936 & 30.09/0.9024 & 30.17/0.9034 & {\color{blue}{30.87}}/{\color{blue}{0.9121}} & {\color{red}{31.16}}/{\color{red}{0.9157}} \\
\hline
\end{tabular}
}

\label{tab:benchmark results}
\end{table*}
\section{EXPERIMENTAL RESULTS}
\subsection{Implementation Details}
In the proposed networks, we set 3$\times$3 as the size of all convolutional layers with one padding and one striding except convolutional layers of local and global feature fusion. The filter size of local and global feature fusion is 1$\times$1 with no padding and one striding. Low-level feature extraction layers and feature fusion layers have 64 filters. The number of RDBs $K$ is 9, and the number of RDGs $G$ is 6. In RD, the number of distilled filters $d$ is 8. We treat the network with 256 original filters in each RDB as DRN+. Other layers in each RDB are followed by the exponential linear unit(ELU~\cite{ELU}) with parameter 0.2. The SR results are evaluated with PSNR and SSIM~\cite{ssim}. We train our model with ADAM optimizer~\cite{ADAM} with MAE loss by setting $\beta_{1}$=0.9, $\beta_{2}$=0.999 and $\epsilon$=$10^{-8}$. The learning rate is $10^{-4}$ and halve at each 200 epochs.We trained DRN and DRN+ for about 800 and 300 epochs respectively.
\subsection{Datasets}
By following many existing image SR methods~\cite{EDSR,MSRN}, we use 800 training images of DIV2K dataset~\cite{DIV2K} as training set and five standard benchmark dataset: Set5~\cite{Set5}, Set14~\cite{Set14}, Urban100~\cite{Urban100}, BSDB100~\cite{B100} and Manga109~\cite{Manga109} as testing set. We set the batchsize to 16. The size of the input image is 48$\times$48.~Instead of transforming the RGB patches into a YCbCr space, we use 3 channels images information from the RGB patches in order to keep the images real.

\subsection{Comparisons with state-of-the-arts}
We compare our method with 10 state-of-the-art methods: A+~\cite{A+},  SRCNN~\cite{SRCNN}, FSRCNN~\cite{FSRCNN}, VDSR~\cite{VDSR}, DRCN~\cite{DRCN}, LapSRN~\cite{LapSRN}, IDN~\cite{IDN}, SRMDNF~\cite{SRMDNF}, EDSR~\cite{EDSR} and MSRN~\cite{MSRN}. We also use self-ensemble~\cite{self-ensemble} to improve our models.

Table \ref{tab:benchmark results} shows quantitative comparison for $\times$2,$\times$3 and $\times$4 SR. Compared with previous methods, our DRN+ performs the best on most datasets with all scaling factors. Even with 64 filters, our DRN is also better than other comparison methods on most datasets.~Table~2 shows ablation test on RDBs. The model with RDBs has a great performance, and DBN+ has a better performance. In Fig.\ref{Cooking}, Fig.\ref{fig:Image027 x3}, Fig.\ref{fig:Image044 x3} and Fig.\ref{fig:Image059 x4} we present visual performance on different datasets with different upscaling factors. Fig.\ref{Cooking} shows visual comparison on scale $\times$4. For image {\em "Image075"}, we observe that most methods can't recover texture on the windows. In contrast, our DRN can better alleviate blurring artifacts and recover details consistent with the Groundtruth. In Fig.\ref{fig:Image027 x3} we observe that the lines of {\em"Image027"} recovered by most methods don't correspond to Groundtruth images well. However, the DRN our proposed  have accurately recoverd the lines. Fig.\ref{fig:Image044 x3} showing {\em"ppt3"}, although most methods have different degrees of blurring on the word "with", the DRN accurately removes the blurs in the picture that people can recognize that the word is "with". In Fig.\ref{fig:Image059 x4} most methods don't recover lines of windows in {\em"Image059"} that inconsistent with Groundtruth image, our DRN accurately restores the lines and removes the blur almost.
%More visual results are provided in supplementary material.

We also compared the trade-offs between performance and network parameters from DRN networks and existing networks. Fig. 2 shows the PSNR performance versus number of parameters, where the results are evaluated with Set5 dataset for 4$\times$ upscaling factor. We can see our DRN network is better than a relatively small models. In addition, the DRN+ achieves higher performance with 54\% fewer parameters compared with EDSR. These comparisons show that our model has a better trade-off between performance and model size.

\section{Conclusion}
In this paper, we propose a simple and efficient distilling with residual network(DRN) for SISR, which is better than most of the state-of-the-art methods and has fewer parameters. Based on resiudal distilling(RD), the DRN inherits the advantages of the dense residue and connection paths, to achieve an effective reuse and re-exploitation. Our DRN and DRN+ have better tradeoff between model size and performance. In the future, we will apply this model to other areas to such as de-raining, dehanzing, and denoising.
% References should be produced using the bibtex program from suitable
% BiBTeX files (here: strings, refs, manuals). The IEEEbib.bst bibliography
% style file from IEEE produces unsorted bibliography list.
% -------------------------------------------------------------------------
\bibliographystyle{IEEEbib}
\bibliography{reference}

\end{document}